\documentclass[11pt]{article}

\usepackage[preprint]{acl}
\usepackage{float}
\usepackage{times}
\usepackage{latexsym}
\usepackage[T1]{fontenc}
\usepackage[utf8]{inputenc}
\usepackage{microtype}
\usepackage{graphicx}
\usepackage{etoolbox}
\usepackage{booktabs}
\usepackage{multirow}
\usepackage{amsthm}
\usepackage{makecell}
\usepackage{amsmath}
\usepackage{subcaption}
\usepackage[most]{tcolorbox}
\usepackage{listings}
\usepackage{enumitem}
\theoremstyle{definition}
\newtheorem{example}{Example}

\makeatletter
\newif\if@isreview
\@ifpackagewith{acl}{review}{\@isreviewtrue}{\@isreviewfalse}
\newcommand{\redact}[1]{%
  \if@isreview\texttt{[redacted for peer review]}\else#1\fi}
\makeatother

\title{Reasoning Primitives in Hybrid and Non-Hybrid LLMs: Do Architectural Differences Yield Advantages in State-Tracking and Recall?}

\author{
  \textbf{Shivam Rawat},
  \textbf{Lucie Flek},
  \textbf{Florian Mai},
\\
  \textbf{Nicholas Kluge Corr\^ea}
\\
  Lamarr Institute for Machine Learning and Artificial Intelligence \\
  Rheinische Friedrich-Wilhelms-Universit\"at Bonn \\
  \small \textbf{Correspondence:} \href{mailto:srawat@uni-bonn.de}{srawat@uni-bonn.de}
}

\begin{document}
\maketitle

\begin{abstract}
Reasoning in large language models is often discussed as a single capability, but some of its gains may stem from simpler underlying operations. We examine two such primitives, recall and state-tracking, through five controlled task families centered on state-based recall, and compare matched transformer and hybrid architectures with and without reasoning augmentation. Across the suite, reasoning-augmented variants substantially outperform instruction-only variants, often by large margins. This pattern is consistent with the \textit{State over Tokens} view: externalized reasoning traces help because they carry the intermediate state forward in token space. By contrast, hybrid inductive bias does not yield a uniform advantage in accuracy once reasoning tokens are available. When architectural differences do appear, they follow task structure: the hybrid Think model is more robust on strictly sequential chained updates, whereas the transformer Think model is more robust on flat multi-hop retrieval. We therefore cast the main contribution of this study as a descriptive account of what drives performance on state-based recall tasks: reasoning-token augmentation appears to be the dominant factor, while hybrid advantages are narrower, task-dependent, and potentially more about inference efficiency than overall capability. We also release the codebase and data required to reproduce these results (Appendix~~\ref{appendix:scientific-artifacts}).
\end{abstract}

\section{Introduction}
\label{sec:intro}

Early work on Chain-of-Thought reasoning established that inducing large language models (LLMs) to generate intermediate steps often improves performance on complex tasks \cite{wei2022chain}. Subsequent work expanded this picture across prompting strategies \cite{yao2023tree}, architectural settings \cite{geiping2025scaling}, and training regimes \cite{wang2025thinking,yang2025qwen3,bakouch2025smollm3,zeng2026glm}. The harder question is why reasoning tokens help. The dominant narrative is that they let the model faithfully explain its own reasoning process, but that account is difficult to test and does not explain why reasoning traces can look coherent without transparently mirroring the underlying computation.

A broad body of literature treats reasoning in LLMs as structured inference via intermediate textual computation \cite{liu2025logical}, but the mechanism underlying the gains remains contested \cite{kambhampati2025reasoning}. Recent work~\cite{levy2025state} offers a useful lens: the \textbf{State over Tokens} (SoT) formulation, which argues that reasoning tokens help because they let models carry intermediate state forward in token space. Across autoregressive generation, internal activations do not persist, but the token prefix does. Intermediate tokens, therefore, become the durable carrier of state across successive forward passes. On this account, reasoning helps not because the model faithfully explains itself, but because extra tokens allow it to externalize partial results, condition on them again, and turn a sequence of bounded computations into a cumulative one. This naturally raises an architectural question: if reasoning gains come from storing state in token space, when should built-in mechanisms for persistent state still matter?

Once the problem is framed in these terms, the architectural question becomes more concrete. Standard softmax-based transformers excel at content-based retrieval but face well-known scaling constraints: full self-attention requires $O(n^2)$ computation and memory, and even with KV caching, sequence length remains a bottleneck as memory grows with stored keys and values \cite{fichtl2025end,li2024survey}. Recurrent and state-space architectures offer more favorable asymptotic profiles, but at the cost of precise content-addressable retrieval \cite{yang2024parallelizing,yang2024gated,dao2024transformers}. This trade-off motivates \textit{hybrid architectures}, which combine both mechanisms \cite{merrill2026olmo}. In that account, attention supports fine-grained recall over static context, whereas recurrent components support a persistent, efficiently updated latent state. The two are best understood as complementary inductive biases rather than as a single performance hierarchy.

These observations suggest that if reasoning tokens improve performance by externalizing state, then the relevant underlying capabilities may not be ``reasoning'' in the abstract but simpler operations that support stateful computation. We therefore study two such primitives, \textbf{Recall} and \textbf{State-tracking} (Appendix \ref{app:primitives-formal}), and ask: once reasoning is decomposed into these computational demands, does performance primarily depend on reasoning augmentation itself, or do architectural differences become visible in systematic ways?

We evaluate matched transformer and hybrid models, with and without reasoning augmentation, on five procedurally generated tasks that probe the same underlying challenge: state-based recall under different surface forms and computational demands. The tasks are not five unrelated benchmarks, but controlled variations on the same problem. Each varies along two axes of difficulty: $m$ (retrieval complexity) and $n$ (state maintenance complexity). Across the suite, reasoning augmentation drives the largest gains, while architectural differences are conditional and become visible only in particular task regimes. When hybrids do have an advantage, it may primarily concern inference efficiency rather than overall accuracy. The contribution of the paper is therefore a descriptive account of when reasoning tokens suffice and when task structure makes architectural differences visible on state-based recall problems.

\section{Methods}
\label{sec:methods}

\subsection{Recall and State-Tracking as Reasoning Primitives}

To make this notion of reasoning operational, we decompose it into two complementary primitives following~\cite{merrill2026olmo}: \textbf{State-tracking}, the maintenance and update of structured variables across sequential transformations, and \textbf{Recall}, the high-precision retrieval of specific information from context (formal definitions in Appendix~\ref{app:primitives-formal}). Their combination yields \textbf{State-based recall}, where the retrieval address is itself produced by a sequence of state updates. In that setting, the task is not reducible to either primitive alone: pure attention lacks stable iterative state evolution, while pure recurrence compresses away fine-grained addressability. This framing lets us describe the task suite in terms of computational demands rather than benchmark labels. Appendix~\ref{app:task-details} provides full task descriptions and examples.

\subsection{Experimental Design}

We evaluate matched transformer and hybrid models, with and without reasoning augmentation, on five procedurally generated tasks that probe the same underlying challenge: state-based recall under different surface forms and computational demands. The tasks are controlled variations of a single problem, each varying along two difficulty axes: $m$ (retrieval complexity) and $n$ (state-maintenance complexity).

\paragraph{Models.} We compare OLMo3-7B~\cite{olmo2025olmo3} and OLMo-Hybrid-7B~\cite{merrill2026olmo}, a matched pair sharing the same data mixture, optimization schedule, and training recipe. For each architecture, we evaluate an instruction-tuned and a reasoning-augmented variant: \textit{OLMo-3-7B-Instruct|Think} and \textit{OLMo-Hybrid-Instruct|Think-SFT-7B}. This design lets us separate the effect of reasoning augmentation from the effect of architectural class. Architecture and inference details are deferred to Appendix~\ref{appendix:infra-eval-details}.

\paragraph{Evaluation.} All tasks are free-form generation scored by exact matching. Difficulty is parameterized by $m$ (retrieval complexity) and $n$ (state maintenance complexity), with 1000 instances per $(m,n)$ bin (500 for Dyck). All tasks require JSON output \texttt{\{"answer": "A|B|C|D"\}}. We report \textit{accuracy} and \textit{parse rate} (the fraction of parseable responses) separately so that generation failures are not conflated with genuine errors.

\paragraph{Tasks.} We evaluate five task families ordered from the minimal baseline to increasingly demanding variants of the same broader problem (Table~\ref{tab:task-descriptions}; details in Appendix~\ref{app:task-details}). \textbf{OLMo Original}~\cite{merrill2026olmo} is the replication baseline. \textbf{Astro Recall} preserves the same pointer-tracking structure with table lookup. \textbf{Dyck} isolates state-tracking. \textbf{Collisions} introduces chained sequential updates, and \textbf{DAG Arithmetic} stresses flat multi-hop retrieval over a growing registry~\cite{motwani2026longcot}. Together, the suite shows how performance changes as the balance between retrieval and state maintenance shifts.

\begin{table}[t]
\centering
\small
\setlength{\tabcolsep}{4pt}
\begin{tabular*}{\columnwidth}{@{\extracolsep{\fill}}p{2.2cm}p{5.0cm}@{}}
\toprule
\textbf{Task} & \textbf{Description} \\
\midrule
OLMo Original & Bit value after pointer swaps in a synthetic array \\
Astro Recall  & Planet name after variable swaps on orbital periods \\
Dyck Language & Correct closing bracket in a masked sequence \\
Collisions    & Final velocity after an elastic collision chain \\
DAG Arithmetic & Variable value after layered arithmetic dependencies \\
\bottomrule
\end{tabular*}
\caption{
Task descriptions and difficulty parameters.
For each task, $m$ controls retrieval complexity and $n$ controls state-maintenance complexity: OLMo Original ($m$ = array length, $n$ = \# swaps), Astro Recall ($m$ = \# table rows, $n$ = \# swaps), Dyck Language ($m$ = stack depth, $n$ = sequence length), Collisions ($m$ = \# particles, $n$ = \# collisions), and DAG Arithmetic ($m$ = \# variables, $n$ = \# layers).
}
\label{tab:task-descriptions}
\end{table}

\section{Results}
\label{sec:results}

The results are most informative when read comparatively across tasks rather than as isolated leaderboards. Two patterns organize the section: reasoning augmentation is the main source of gains, and architectural differences, when they appear, track task structure rather than difficulty alone. We therefore present the tasks from the OLMo baseline to settings that place increasingly distinct demands on state maintenance and retrieval.

\paragraph{OLMo Original.} The replication baseline already establishes the main pattern. The transformer Instruct model consistently exceeds the hybrid Instruct model (0.43--0.53 vs.\ 0.32--0.43), but once reasoning augmentation is introduced, the two Think variants start near ceiling at $(4,4)$ and decline in parallel toward chance by $(64,64)$, without a consistent architectural difference in accuracy. Transformer variants also maintain higher parse rates, though this does not translate into an accuracy advantage at the Think level (Appendix~\ref{app:additional-figures}, Figures~\ref{fig:olmo_acc},~\ref{fig:olmo_parse}).

\paragraph{Astro Recall.} A more naturalistic retrieval surface leaves the overall story intact. Both Instruct variants remain near chance across the full range while maintaining high parse rates. Both Think variants perform strongly at low difficulty ($\approx$0.95--0.97 at $(4,4)$) but, by $(32,32)$, increasingly exhaust the inference budget before producing a parseable answer, with the hybrid Think model reaching near-zero parse rate earlier (Appendix~\ref{app:additional-figures}, Figures~\ref{fig:astro_acc},~\ref{fig:astro_parse}). This reinforces the importance of reasoning augmentation while also showing that reasoning tokens alone do not eliminate failure once both retrieval and state-maintenance demands become large.

\paragraph{Dyck Language.} Dyck isolates the state-tracking side of the decomposition and provides the clearest case in which architecture may matter. Both Instruct variants stay close to the 0.33 random baseline. Among Think variants, \texttt{OLMo-Hybrid-Think-SFT-7B} attains the strongest results in the paper (peak 0.79 at $(2,128)$ and never below 0.38), whereas \texttt{OLMo-3-7B-Think} is more variable (peak 0.67 and 0.22 at $(4,2048)$); the hybrid Think model's parse rate declines at high difficulty even when accuracy remains comparatively strong (Appendix~\ref{app:additional-figures}, Figures~\ref{fig:dyck_acc},~\ref{fig:dyck_parse}). The pattern is consistent with a possible hybrid advantage in pure state-tracking, but we treat it as suggestive rather than general.

\begin{figure}[t]
\centering
\includegraphics[width=\columnwidth]{"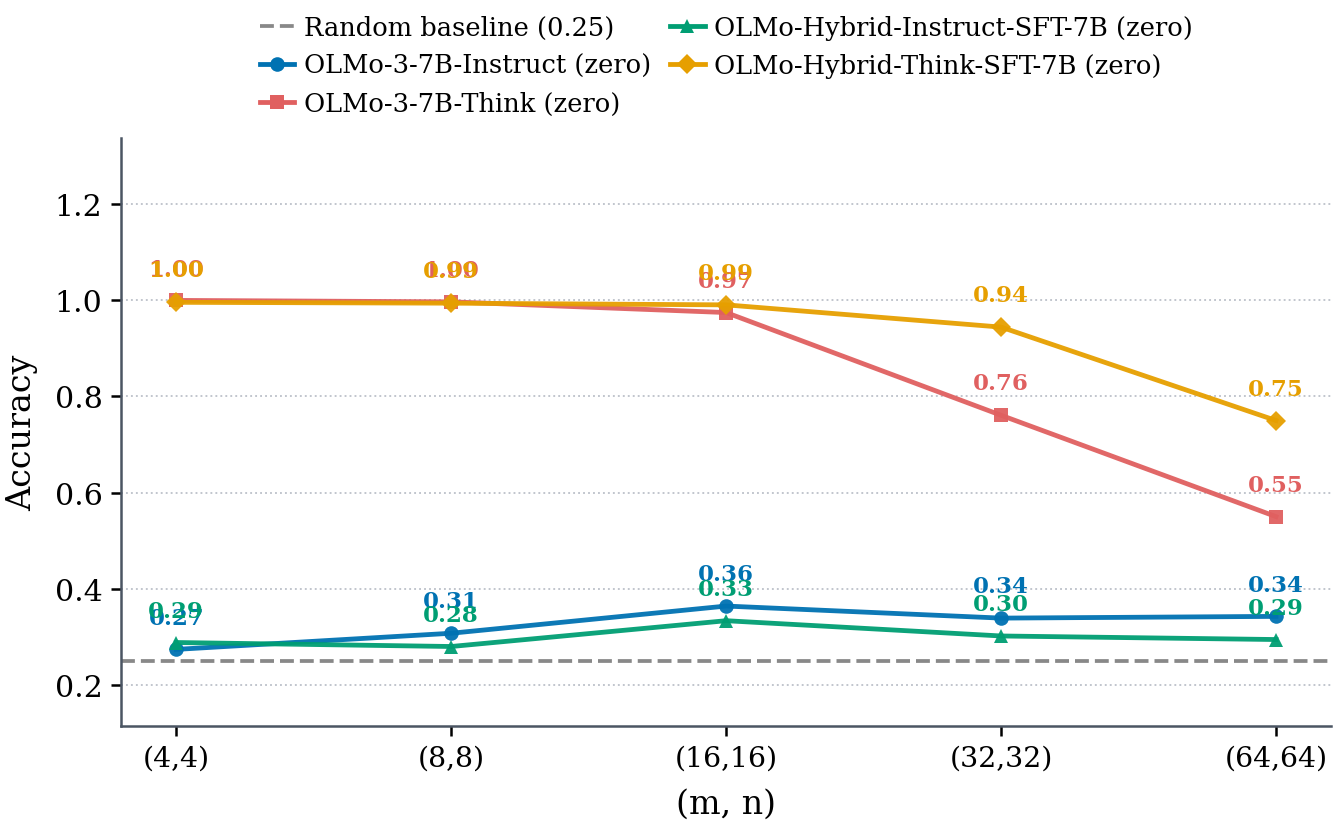"}
\caption{Collision Simulator accuracy.}
\label{fig:collisions_acc}
\end{figure}

\begin{figure}[t]
\centering
\includegraphics[width=\columnwidth]{"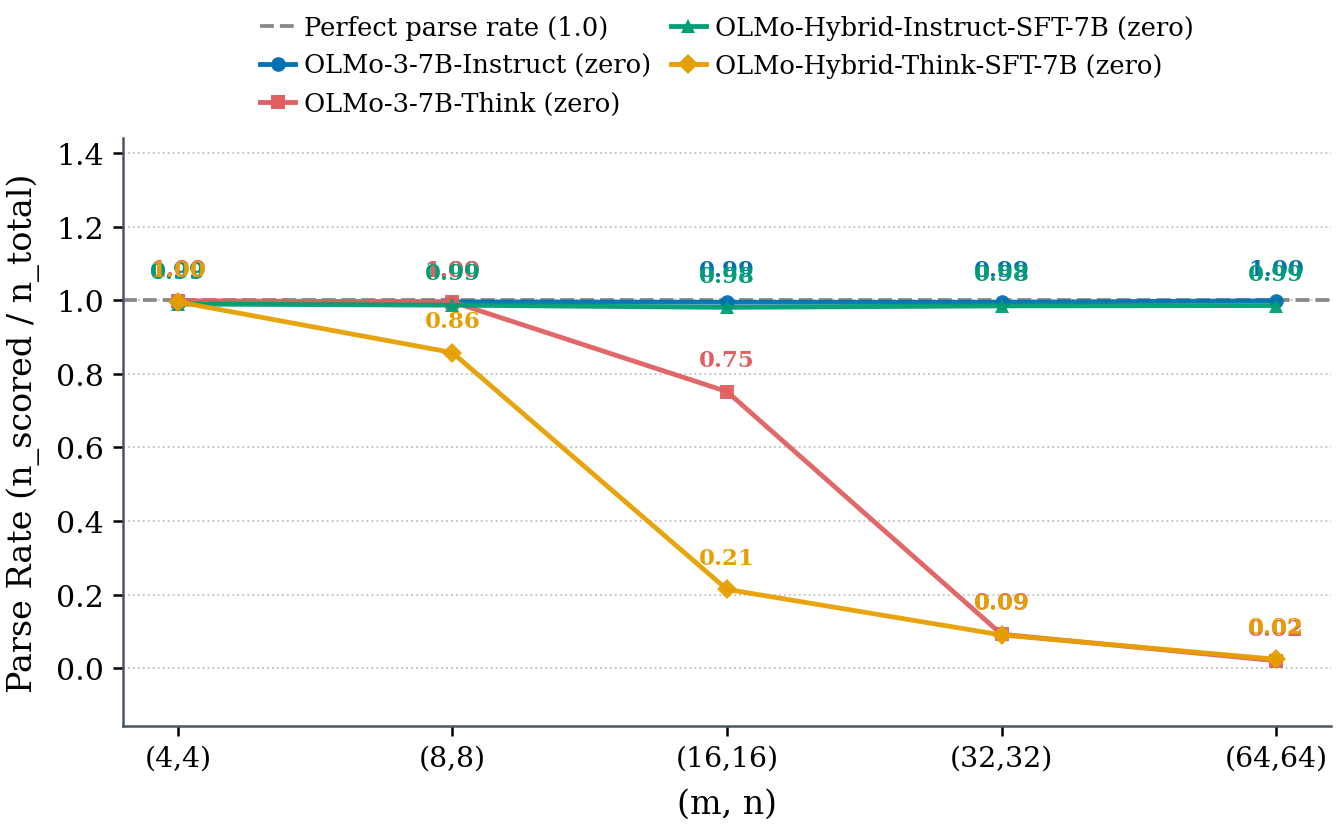"}
\caption{Collision Simulator parse rate.}
\label{fig:collisions_parse}
\end{figure}

\paragraph{Collision Simulator.} Collision Simulator adds back state-based recall with strictly chained sequential updates, making it the strongest test of whether architectural differences follow task structure. Both Instruct variants remain near chance (0.27--0.40). At low difficulty both Think variants are near ceiling, but beyond $(16,16)$ they separate: \texttt{OLMo-3-7B-Think} reaches 0.49 at $(32,32)$ and increasingly exhausts the inference budget before producing a parseable final answer at $(64,64)$, whereas \texttt{OLMo-Hybrid-Think-SFT-7B} retains 0.94 and 0.75 at the same settings (Figures~\ref{fig:collisions_acc},~\ref{fig:collisions_parse}). This is the strongest evidence in the main results that hybrids can help when success depends on sustaining a long chain of sequential state updates.

\begin{figure}[t]
\centering
\includegraphics[width=\columnwidth]{"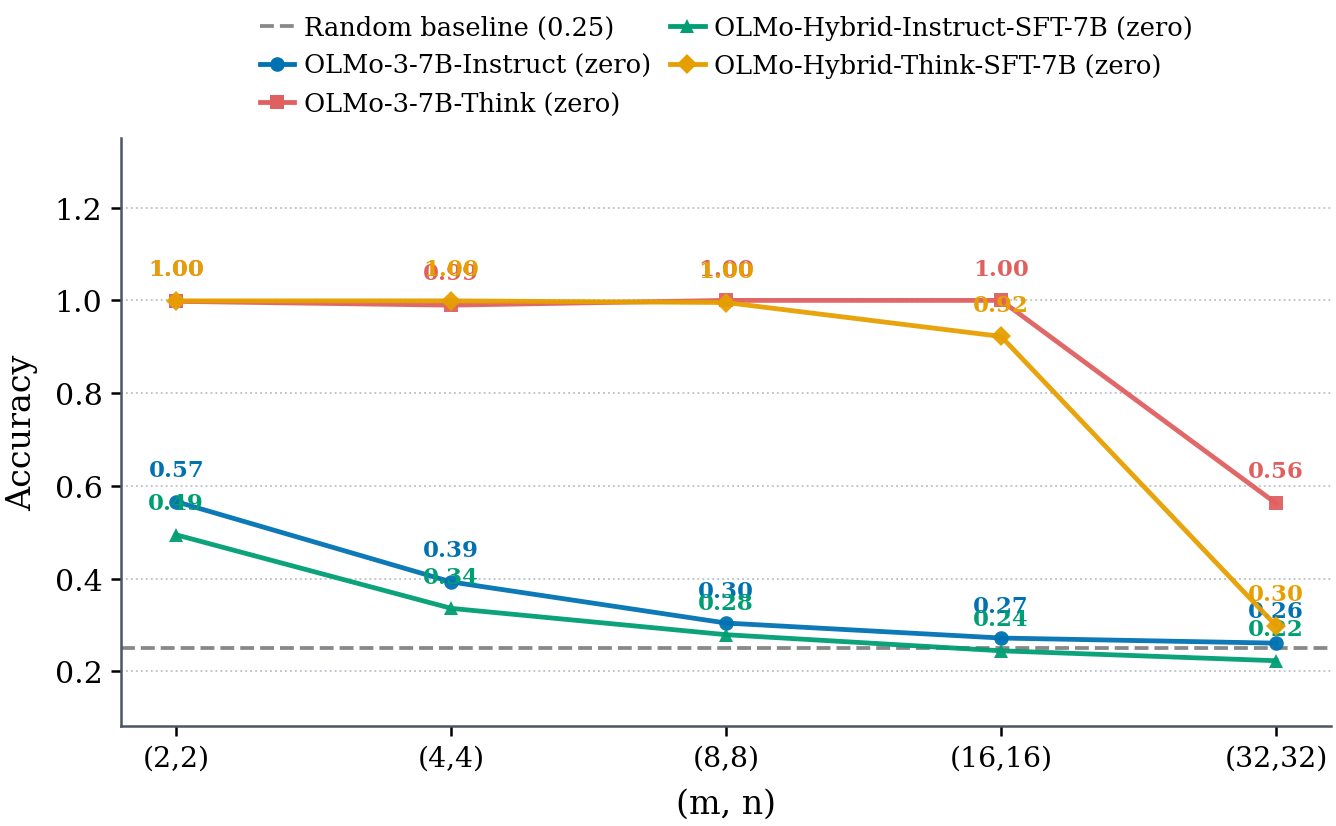"}
\caption{DAG Arithmetic accuracy.}
\label{fig:dag_acc}
\end{figure}

\begin{figure}[t]
\centering
\includegraphics[width=\columnwidth]{"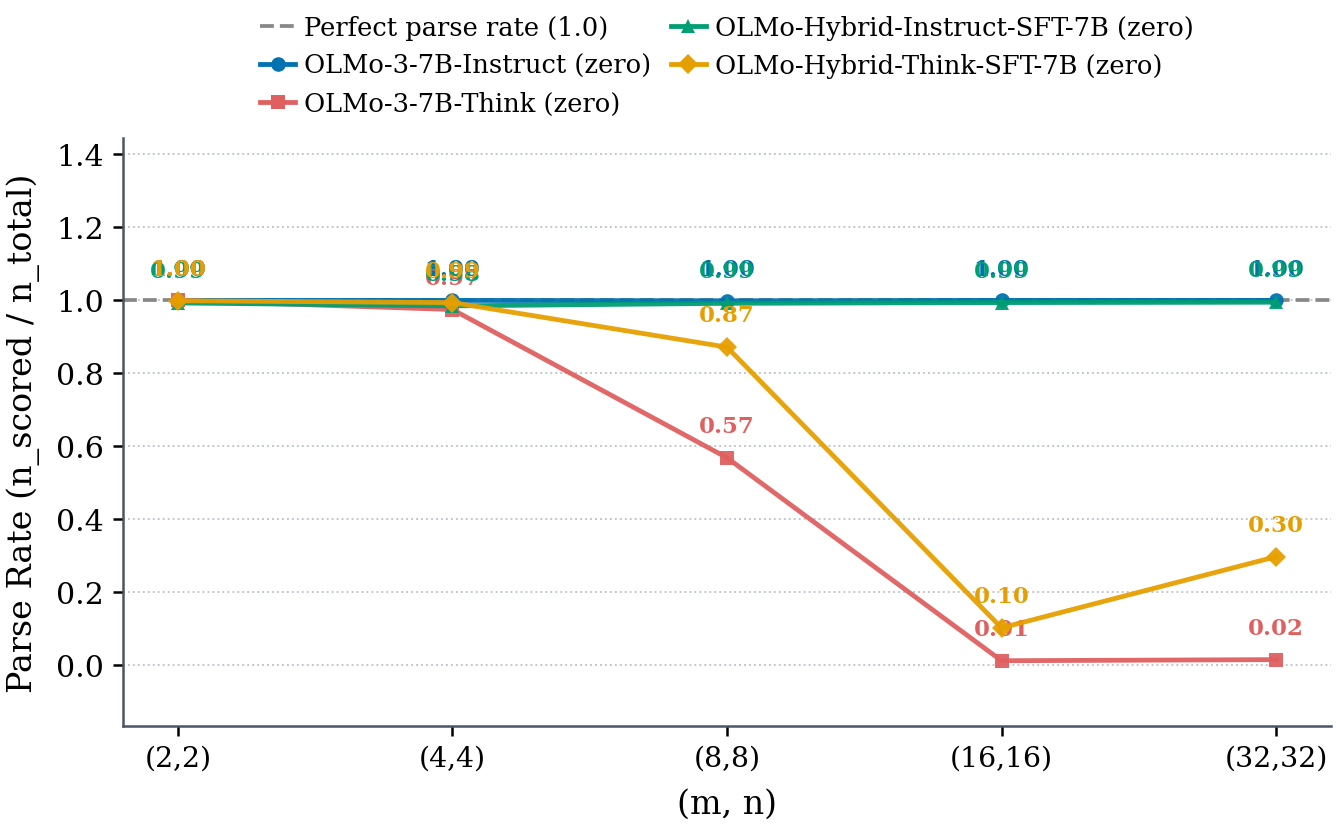"}
\caption{DAG Arithmetic parse rate.}
\label{fig:dag_parse}
\end{figure}

\paragraph{DAG Arithmetic.} DAG Arithmetic stresses a different part of the state-based recall space: flat multi-hop retrieval over an expanding registry rather than long sequential overwriting. The Instruct variants begin above chance at $(2,2)$ and approach the 0.25 baseline by $(32,32)$ with little separation. The Think variants are both perfect from $(2,2)$ through $(8,8)$, then diverge in the opposite direction to collisions: at $(32,32)$ \texttt{OLMo-3-7B-Think} retains 0.56 while \texttt{OLMo-Hybrid-Think-SFT-7B} reaches 0.30, even though the hybrid Think model maintains the higher parse rate throughout (Figures~\ref{fig:dag_acc},~\ref{fig:dag_parse}). This reversal argues against any single architectural ranking and instead supports a task-dependent interpretation, with transformers appearing more robust when the challenge is content-addressable multi-hop retrieval.

\section{Discussion \& Conclusion}
\label{sec:discussion}

Across the five tasks, two claims are most directly supported by the data. First, reasoning augmentation is the dominant driver of performance gains: Think variants consistently outperform Instruct variants, often by large margins. This pattern is consistent with the SoT interpretation, in which externalized traces help by carrying the intermediate state in token space. Because the task suite was designed to stress recall and state-tracking, the results also suggest that these primitives capture part of what current ``reasoning'' improvements rely on.

Second, the advantages of hybrids are conditional rather than general. On the OLMo Original baseline, the two Think models are effectively matched in accuracy, and across the suite, no uniform hybrid advantage emerges. Dyck suggests a possible hybrid benefit in pure state-tracking, while Collisions and DAG Arithmetic reveal opposite advantages depending on whether the task emphasizes sequential state propagation or flat multi-hop retrieval. This pattern is more consistent with task-dependent inductive biases than with a single architectural hierarchy.

The broader implication is that hybrids may matter most for efficiency rather than for uniformly higher capability. Long-context reasoning places strain on KV-cache growth, memory bandwidth, and throughput, all of which scale poorly under full quadratic attention. By mixing sparse full-attention layers with recurrent-style state compression, hybrids can reduce prefilling and decoding costs while preserving enough global interaction to remain competitive on complex tasks. Our contribution is therefore both empirical and methodological: we provide a controlled comparison of transformer and hybrid reasoning models, and we introduce a compact suite that tests architectural claims using recall and state tracking rather than broad benchmark aggregates. Taken together, the results suggest that the main benefits of hybrid models may lie less in qualitatively different reasoning abilities than in more efficient support for those abilities at scale.

\section{Limitations}
\label{sec:limitations}

This study has several limitations that the authors acknowledge. First, the architectural comparison is restricted to a single matched 7B model pair, which makes the comparison relatively controlled and fair, but also limits generalization: the observed patterns may be specific to the OLMo3 family, this particular scale, or the associated training recipe and hybrid design choices.\footnote{Although, we were unable to find hybrid non-hybrid models that would allow for this level of fairness in terms of comparison, i.e., models mainly differ in terms of arquitecture, but nothing else.} Second, all evaluations are conducted on synthetic tasks. This choice is intentional, since the benchmark is designed to isolate recall, state-tracking, and their composition into state-based recall, but it necessarily narrows the scope of the claims that can be made. The tasks should therefore be interpreted as probes of computational structure rather than as direct measurements of broad reasoning ability, and the present results do not establish that the same trends will necessarily transfer to more naturalistic domains such as long-document question answering, mathematical reasoning, or tool-use settings. In addition, the experiments were conducted under a fixed inference budget. Although most failed Think trajectories appear to enter repetitive ``doom loops'' rather than making measurable progress toward a solution, it remains possible that different budget regimes could alter some of the observed comparisons. A more systematic analysis of budget scaling would help clarify whether the main differences observed here reflect representational limitations or simply the increasing computational cost of sustaining long reasoning traces. Finally, the paper evaluates token-space reasoning only indirectly. While the overall empirical pattern is broadly consistent with the SoT interpretation, we do not directly compare natural-language reasoning traces against compressed traces, nor do we evaluate architectures that partially externalize reasoning into latent representations. Such comparisons remain important directions for future work to disentangle the relative contributions of state externalization, architectural design, and inference efficiency to the reasoning gains observed in this work.

\section{Ethics Statement}

This research was conducted using publicly available models, and did not involve human subjects or personally identifiable information. The computational resources used were provided by the authors' institution, and the environmental impact of training and inference was minimized to the best of our abilities, while the total cost (i.e., energy consumption and carbon emissions equivalents) has been reported. The findings of this research are intended to advance understanding of reasoning capabilities in language models and do not have direct applications that could be harmful. However, as with all research in artificial intelligence, there is a potential for misuse, and we encourage responsible use of these findings in the development of future AI systems.

\section*{Acknowledgments}

\redact{This research was supported by the state of North Rhine-Westphalia as part of the Lamarr Institute for Machine Learning and Artificial Intelligence. The authors also gratefully acknowledge access to the Marvin cluster at the University of Bonn, along with support from the High Performance Computing Team.}

\bibliography{references}

\appendix

\section{Scientific Artifacts}
\label{appendix:scientific-artifacts}

This study releases the source code and data needed to reproduce the results. The codebase is available at \redact{\href{https://github.com/ultor1996/reasoning_primitives}{github.com/ultor1996/reasoning\_primitives}}.

\section{Formal Definitions of Reasoning Primitives}
\label{app:primitives-formal}

This appendix provides formal definitions of the two reasoning primitives introduced in Section~\ref{sec:methods}.

\paragraph{State-tracking.} State-tracking is the ability to maintain and update structured variables over time under sequential transformations. The critical requirement is not merely the storage of initial values, but the preservation of the evolving relational configuration among variables as successive updates are applied. Formally, letting $S_t$ denote the state at step $t$ and $U_t$ the update function applied at that step, the model must compute:
\begin{equation}
    S_t = U_t(S_{t-1})
    \label{eq:state-tracking}
\end{equation}
and ensure that $S_t$ accurately reflects the cumulative effect of all updates up to time $t$. Recurrent and state-space models are well-matched to this regime because they maintain a compact hidden state that is updated incrementally at each step. Standard attention mechanisms are often less naturally aligned with problems whose difficulty lies in preserving the result of many compositional updates, especially when those updates are long-range, indirect, or involve repeated reassignment.

\paragraph{Recall.} Recall is the ability to retrieve a specific piece of information from a large context with high precision, often when the relevant signal is sparse relative to the total input. Formally, given a context $C$ of $N$ items and a query $q$, the model must compute:
\begin{equation}
    R = \mathrm{Attention}(q, C)
    \label{eq:recall}
\end{equation}
such that $R$ accurately reflects the relevant item in $C$ corresponding to $q$, especially as $N$ grows large and the relevant signal becomes sparser. This is the regime in which softmax attention is most naturally interpreted as a differentiable associative memory mechanism. Purely recurrent or linear state-space models must compress history into a fixed-dimensional state, making fine-grained retrieval progressively harder over long horizons.

\paragraph{State-based recall.} When the two primitives are composed, the retrieval query is not a fixed external input but is itself the output of a state-tracking process. The model must first compute the final state $S_T$ via Equation~\ref{eq:state-tracking} and then use that state as the query for retrieval via Equation~\ref{eq:recall}:
\begin{equation}
    R = \mathrm{Attention}(S_T,\, C)
    \label{eq:state-based-recall}
\end{equation}
This composition makes the task irreducible to either primitive in isolation: an error in state propagation corrupts the retrieval address, and a failure in retrieval produces the wrong answer even given a correct address. Hybrid architectures have been hypothesized to be more expressive due to their inherent architectural design \citep{merrill2026olmo}, which combines the inductive biases.
 
\section{Additional Results}
\label{app:additional-figures}
 
\begin{figure}[H]
\centering
\includegraphics[width=\columnwidth]{"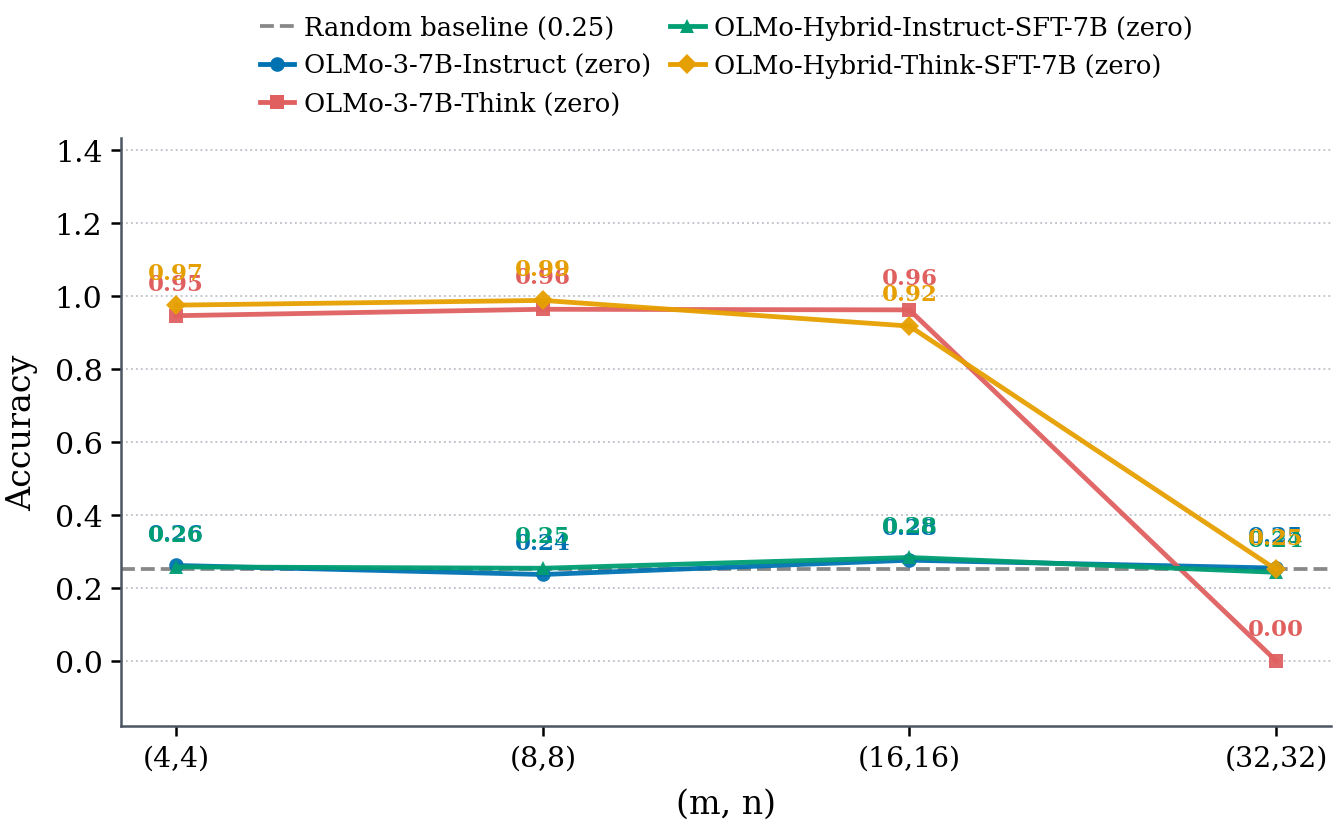"}
\caption{Astro Recall accuracy.}
\label{fig:astro_acc}
\end{figure}

\begin{figure}[H]
\centering
\includegraphics[width=\columnwidth]{"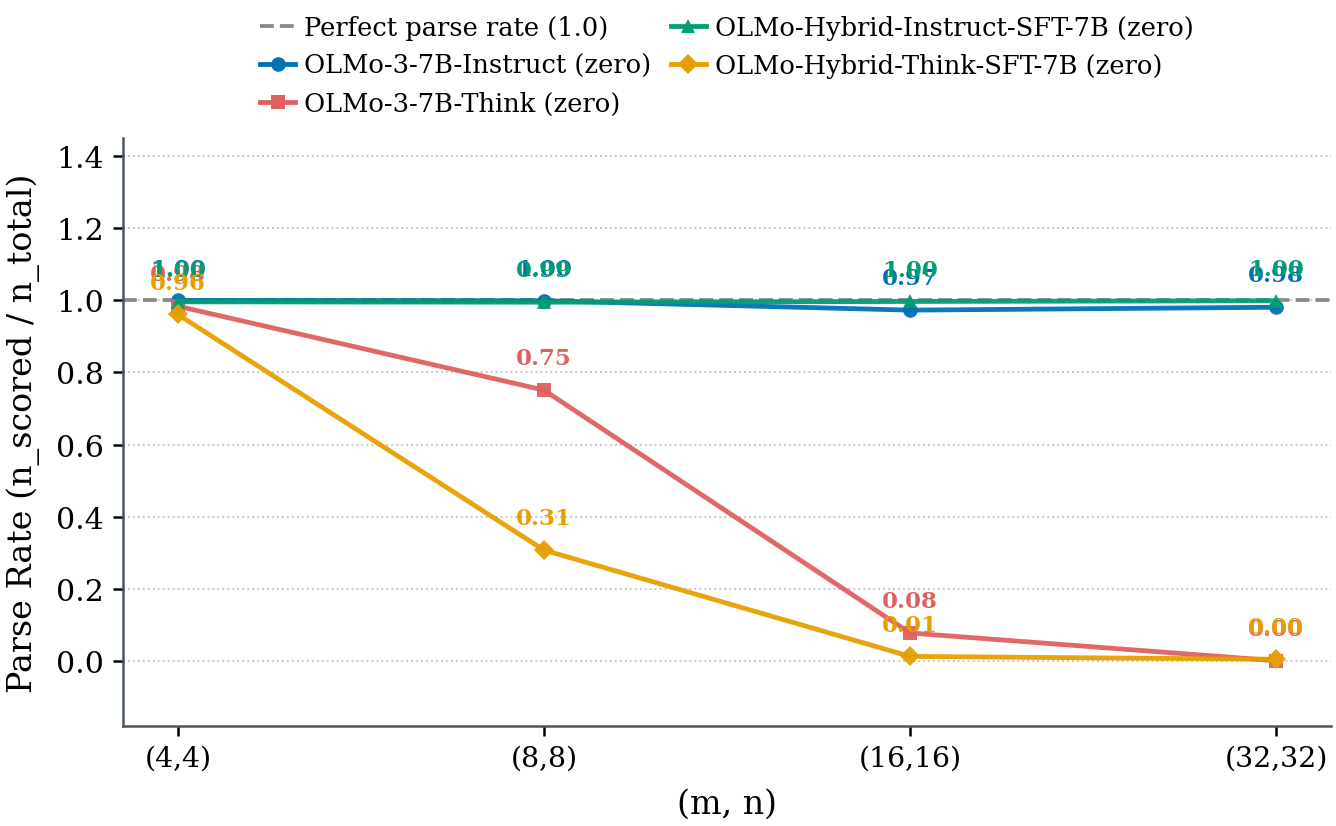"}
\caption{Astro Recall parse rate.}
\label{fig:astro_parse}
\end{figure}

\begin{figure}[H]
\centering
\includegraphics[width=\columnwidth]{"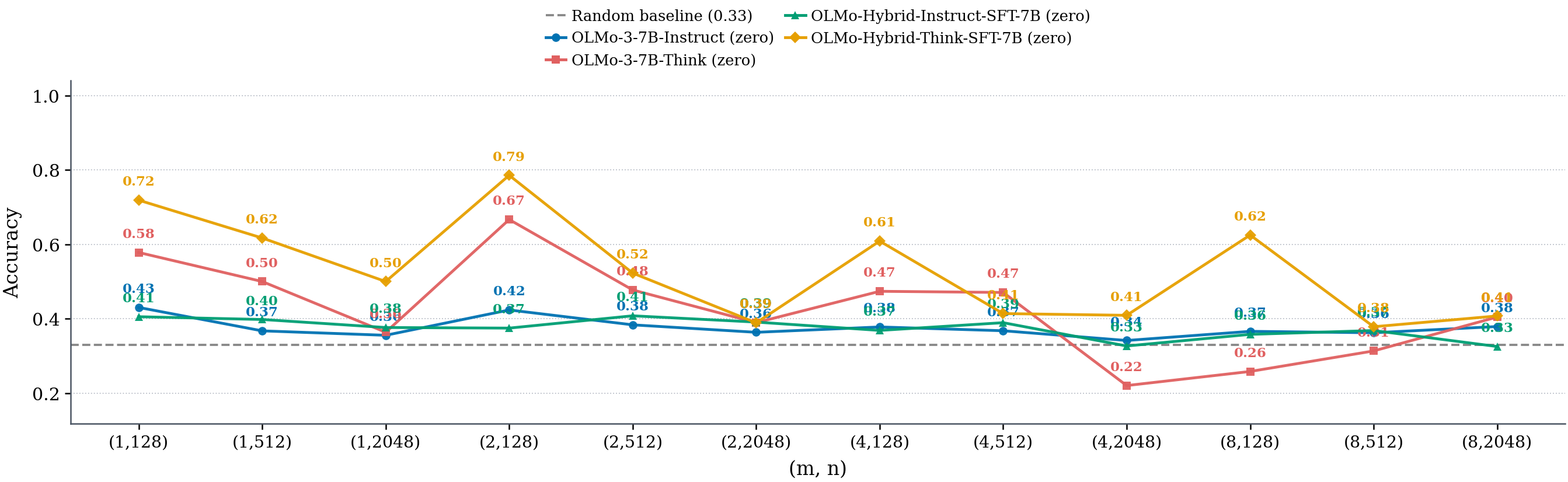"}
\caption{Dyck accuracy.}
\label{fig:dyck_acc}
\end{figure}

\begin{figure}[H]
\centering
\includegraphics[width=\columnwidth]{"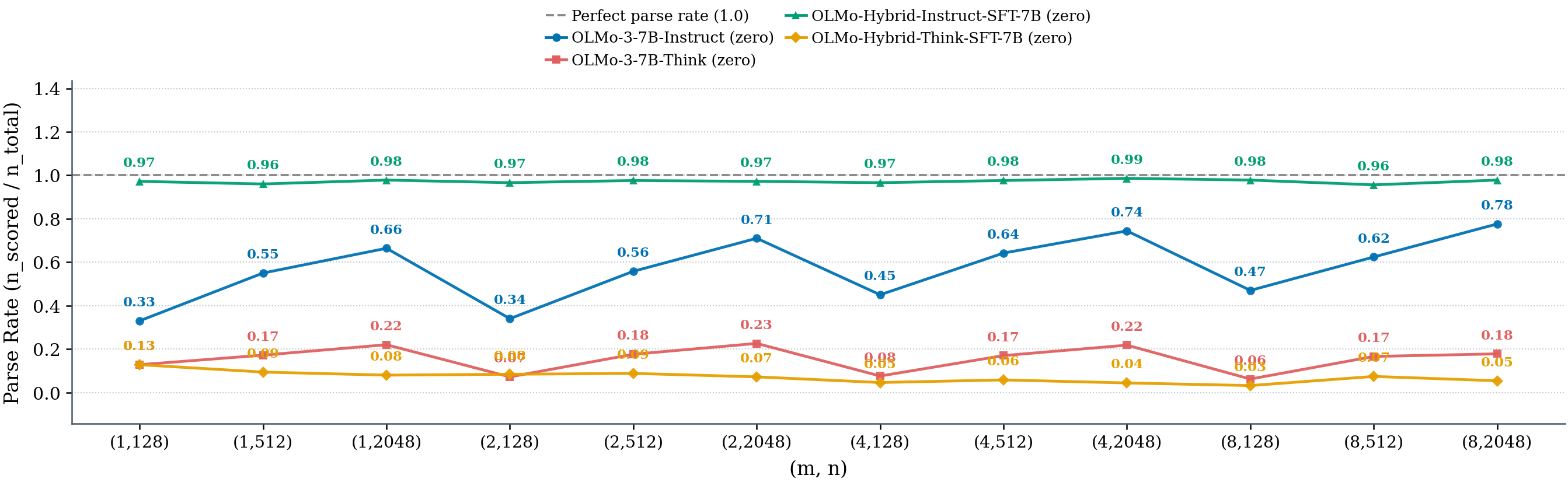"}
\caption{Dyck parse rate.}
\label{fig:dyck_parse}
\end{figure}

\begin{figure}[H]
\centering
\includegraphics[width=\columnwidth]{"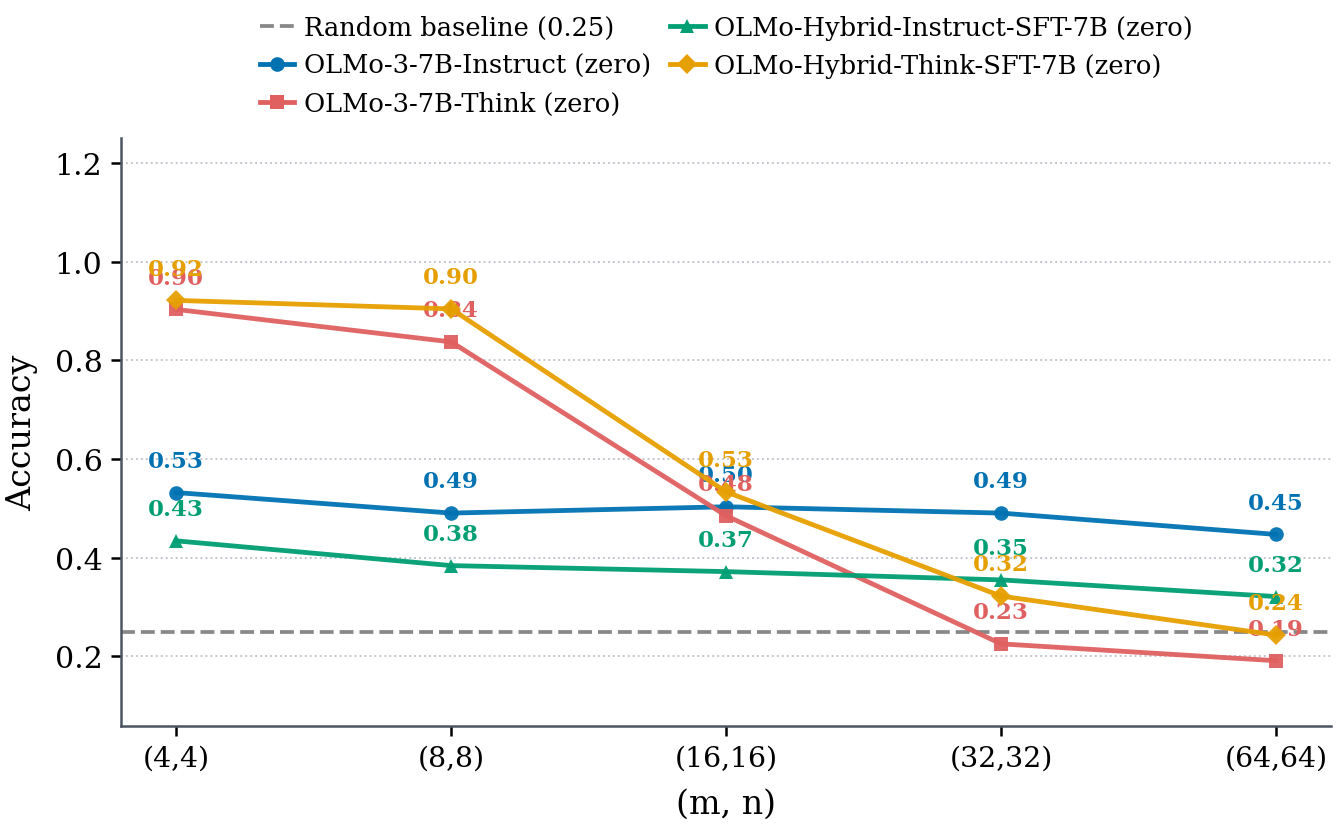"}
\caption{OLMo Original accuracy.}
\label{fig:olmo_acc}
\end{figure}

\begin{figure}[H]
\centering
\includegraphics[width=\columnwidth]{"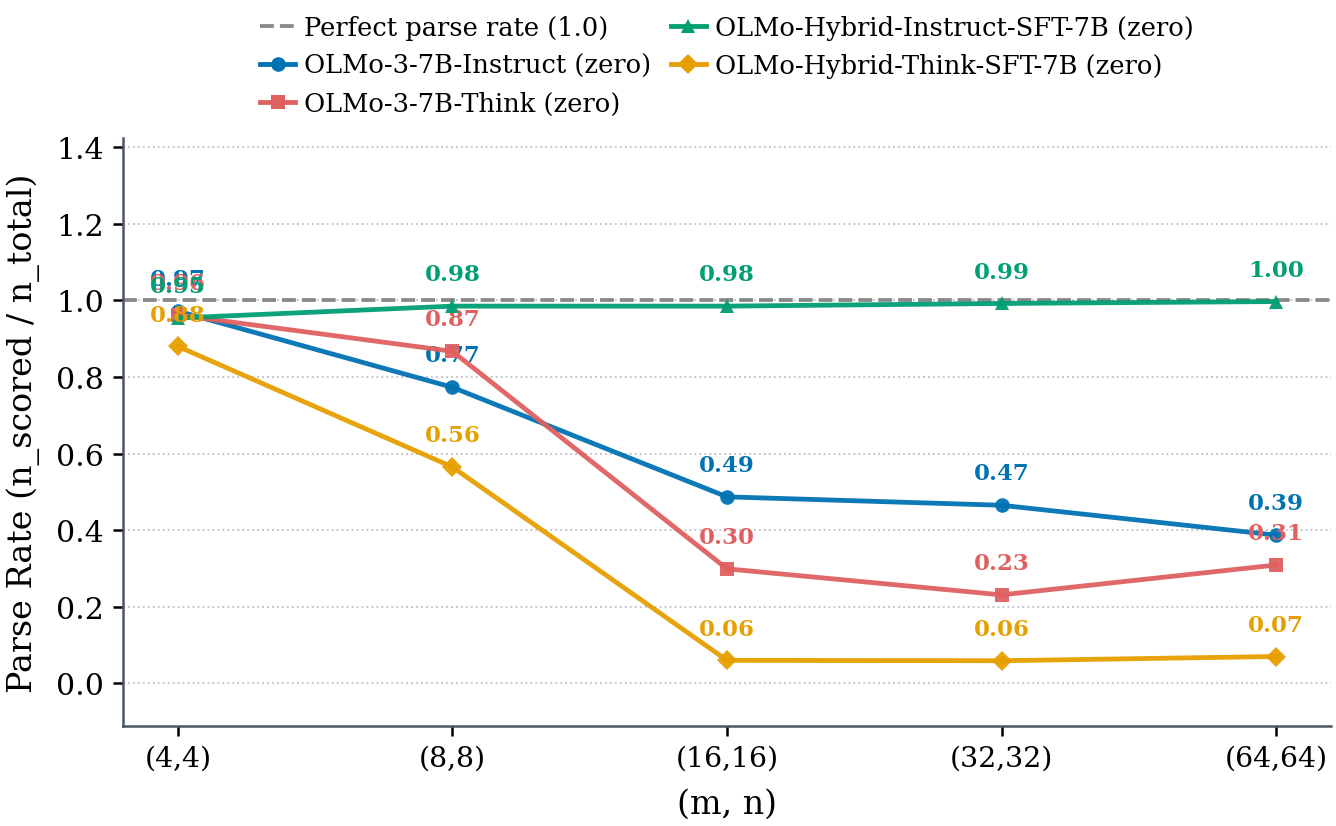"}
\caption{OLMo Original parse rate.}
\label{fig:olmo_parse}
\end{figure}

Figures~\ref{fig:astro_acc}--\ref{fig:olmo_parse} show accuracy and parse rate results for the three task families not shown in the main body: State-based Astro Recall, Dyck Language, and OLMo Original.
 
\section{Task Suite Details and Prompt Examples}
\label{app:task-details}

This appendix groups the task descriptions and prompt templates for the five task families discussed in the main text.

\subsection{Reasoning Primitive Examples}

Before turning to the task families themselves, we first give three short canonical examples illustrating the primitives discussed in Section~\ref{sec:methods}: state-tracking, recall, and their composition into state-based recall, as proposed by \citet{merrill2026olmo}.
 
\begin{example}
\label{ex:state-tracking}
\textbf{State-Tracking}  A canonical state-tracking task requires propagating variable bindings across a chain of dependent updates. The variables are initialized with binary values, then updated through a sequence of $n$ dependent swap operations; the final assertion depends on the entire chain of updates. The challenge lies not in retrieving a fixed fact, but in preserving the result of many compositional updates---precisely the regime where recurrent architectures that maintain an incrementally updated hidden state are well matched:

\begin{tcolorbox}[colback=gray!5!white, colframe=gray!75!black, title=Toxicity Evaluation Prompt, breakable]
\small
\begin{verbatim}
a, b, c, d, e = [0, 1, 0, 0, 1]
a, c = c, e
...
assert a == _
\end{verbatim}

\end{tcolorbox}

\end{example}
 
\begin{example}
\label{ex:recall}
\textbf{Recall}  A canonical recall task requires retrieving a specific item from a large array-like context given a precise index. The variable \texttt{a} holds an index into a long list of bits; the assertion depends solely on retrieving the bit at that position. The challenge is to localize the relevant item without losing fidelity as context length grows---the regime where softmax attention functions as a differentiable associative memory mechanism, and where purely recurrent models that compress history into a fixed-dimensional state face increasing difficulty:

\begin{tcolorbox}[colback=gray!5!white, colframe=gray!75!black, title=Toxicity Evaluation Prompt, breakable]
\small

\begin{verbatim}
bits = [0, 1, 0, 0, ...]
a = 42
assert bits[a] == _
\end{verbatim}

\end{tcolorbox}
\end{example}
 
\begin{example}
\label{ex:state-based-recall}
\textbf{State-Based Recall} A canonical state-based recall task combines both primitives. A long list of bits serves as the memory, and the variables \texttt{a}--\texttt{e} are initialized as indices into that list. After $n$ dependent swap operations, the final assertion depends on both correctly tracking the transformed index (state-tracking) and retrieving the bit at that position (recall). The retrieval step is conditioned on the outcome of the state-tracking step, making the task irreducible to either primitive in isolation:

\begin{tcolorbox}[colback=gray!5!white, colframe=gray!75!black, title=Toxicity Evaluation Prompt, breakable]
\small

\begin{verbatim}
bits = [0, 1, 0, 0, ...]
a, b, c, d, e = 36, 23, 12, 2, 56
a, c = c, e
...
assert bits[a] == _
\end{verbatim}

\end{tcolorbox}
\end{example}

\subsection{OLMo Original (\textit{olmo\_original})}

A synthetic state-based recall task imported directly from ~\cite{merrill2026olmo}. A bit array of length $m$ is generated, and $m$ pointer variables are each initialized to a distinct random index. After $n$ simultaneous swap operations, the model must evaluate \texttt{bits[a]}---combining pointer tracking with binary value lookup. This is the most minimal state-based recall formulation in the suite and serves as the replication baseline.

\begin{tcolorbox}[colback=gray!5!white, colframe=gray!75!black, title=System Prompt, breakable]
\small
You are a strict code-execution engine.

\textbf{Task:}
\begin{enumerate}
\item You are given a bit array and pointer variables (a, b, c, ...).
\item You are given a sequence of simultaneous swap assignments.
\item You must track the pointer values through every swap and then look up the correct bit in the array.
\end{enumerate}

\textbf{Rules:}
\begin{itemize}
\item Each swap line uses Python simultaneous assignment: \texttt{x, y = y, x}.
\item Apply every swap in order; do \textbf{not} skip any.
\item After all swaps, evaluate \texttt{bits[⟨queried variable⟩]}.
\end{itemize}

\textbf{Output requirements:}
\begin{itemize}
\item Return EXACTLY one JSON object, no other text.
\item Example Output: \texttt{\{"answer": "A | B | C | D"\}}
\end{itemize}

\end{tcolorbox}

\begin{tcolorbox}[colback=gray!5!white, colframe=gray!75!black, title=Example Prompt, breakable]
\small
\begin{itemize}
\item \texttt{bits = [0, 1, 1, 0]} \quad (4 bits)
\item \texttt{a, b, c, d = 2, 0, 3, 1} \quad (indices 0..3)
\item Swaps:
\begin{itemize}
\item \texttt{a, b = b, a}
\item \texttt{b, c = c, b}
\end{itemize}
\item Then evaluate: \texttt{bits[a]} (expected 0 or 1)
\end{itemize}

\textbf{Options}
\begin{itemize}
\item[A)] 1
\item[B)] 0
\item[C)] 2
\item[D)] 3
\end{itemize}
\end{tcolorbox}

\subsection{Dyck Language (\textit{dyck})}

The model is given a bracket sequence drawn from the pairs \texttt{()}, \texttt{[]}, and \texttt{\{\}}, with one closing token masked, and must identify the correct closing token by maintaining a stack. Difficulty is controlled by $m$ (stack depth at the query position) and $n$ (total sequence length). This is the only task in the suite containing no retrieval component---it stresses pure state-tracking in isolation.

\begin{tcolorbox}[colback=gray!5!white, colframe=gray!75!black, title=System Prompt, breakable]
\small
You are a strict language validator for Dyck expressions.

\textbf{Rules:}
\begin{itemize}
  \item A Dyck expression uses bracket pairs: \((\,)\), \([\ ]\), \(\{\ \}\).
  \item Every opening bracket must be closed by its exact matching closer.
  \item Brackets must be closed in the correct order (last opened = first closed).
\end{itemize}

\textbf{Task:}
\begin{enumerate}
  \item You are given a Dyck expression with one token masked as \texttt{\_}.
  \item You must determine which token \texttt{\_} must be to keep the expression valid.
\end{enumerate}

\textbf{How to solve:}
\begin{itemize}
  \item Read the sequence token by token from left to right.
  \item Maintain a stack: push every opener \((, [, \{\) onto the stack.
  \item When you see a closer \(), ], \}\), pop the top of the stack and require a matching pair.
  \item The masked token \texttt{\_} must match the opener currently on top of the stack at that position.
  \item Do NOT skip any token. Do NOT guess based on surrounding tokens alone.
\end{itemize}

\textbf{Output requirements:}
\begin{itemize}
  \item Return EXACTLY one JSON object, no other text.
  \item Example Output: \texttt{\{"answer": "A | B | C"\}}
\end{itemize}

\end{tcolorbox}

\begin{tcolorbox}[colback=gray!5!white, colframe=gray!75!black, title=Example Prompt, breakable]
\small
Expression: \texttt{( [ \{ \_ ] ) ( )}

Bracket pairs: \((\,)\) \quad \([\ ]\) \quad \(\{\ \}\). \\
Every opener must be closed by its matching closer in the correct order.

What token must replace \texttt{\_} at the indicated position?

\textbf{Options}
\begin{itemize}
  \item[A)] \texttt{)}
  \item[B)] \texttt{\}}
  \item[C)] \texttt{]}
\end{itemize}
\end{tcolorbox}

\subsection{State-based Astro Recall (\textit{astro})}

Each instance presents the model with a Markdown table of $m$ real exoplanet rows. The data is taken from NASA Exoplanet Archive ~\cite{christiansen2025nasa}. Named variables are initialized to the orbital period values of those rows, then subjected to $n$ simultaneous swap operations. The model must determine which planet name corresponds to the queried variable after all swaps. This couples state-tracking (pointer propagation through swaps) with recall (lookup in a semantically structured multi-attribute table) and tests whether the state-based recall challenge persists when embedded in a more naturalistic retrieval context.

\begin{tcolorbox}[colback=gray!5!white, colframe=gray!75!black, title=System Prompt, breakable]
\small

You are a precise reasoning assistant.

You will be given:
\begin{enumerate}
\item A table of exoplanet data with planet names and various properties.
\item Variable assignments mapping variable names to values from a specific column.
\item One or more swap operations (Python-style simultaneous assignment).
\item A question asking which planet corresponds to a variable after all swaps.
\end{enumerate}

\textbf{How to solve:}
\begin{itemize}
\item Each variable ((a,b,c,\ldots)) starts assigned to a specific value from the table and therefore corresponds to a specific planet.
\item Each swap line exchanges the values of two variables simultaneously: (\texttt{x, y = y, x}) means (x) gets (y)'s current value and (y) gets (x)'s current value.
\item Track which value each variable holds after every swap.
\item At the end, find which planet in the table has the value that the queried variable currently holds.
\item Apply every swap in order. Do NOT skip any.
\end{itemize}

\textbf{Output requirements:}
\begin{itemize}
\item Return EXACTLY one JSON object.
\item No extra text.
\end{itemize}

\textbf{Format:} \texttt{\{"answer": "A | B | C | D"\}}
\end{tcolorbox}

\begin{tcolorbox}[colback=gray!5!white, colframe=gray!75!black, title=Example Prompt, breakable]
\small
\begin{tabular}{@{}l r@{}}
\textbf{Planet} & \textbf{Orbital Period (days)} \\ \hline
Kepler-1 & 2.684 \\
Kepler-2 & 8.798 \\
Kepler-3 & 5.508 \\
Kepler-4 & 2.644 \\
\end{tabular}

\vspace{6pt}
Consider the following Orbital Period (days):
\begin{center}
\texttt{a, b, c, d = 2.684, 8.798, 5.508, 2.644}
\end{center}

Consider the following swapping:
\begin{itemize}
\item \texttt{a, b = b, a}
\item \texttt{b, c = c, b}
\end{itemize}

The Planet with the Orbital Period (days) = \texttt{a} is:

\textbf{Options}
\begin{itemize}
\item[A)] Kepler-2
\item[B)] Kepler-1
\item[C)] Kepler-3
\item[D)] Kepler-4
\end{itemize}
\end{tcolorbox}

\subsection{Collision Simulator (\textit{collisions})}

Each instance places $m$ particles in a one-dimensional system with distinct initial velocities drawn from a large integer pool. A sequence of $n$ pairwise elastic collisions is applied in order; equal-mass particles exchange velocities on contact. Unlike the simultaneous swaps in the astro task, each collision here conditions all subsequent ones, creating a chain of compounding sequential dependencies. This makes the task more adversarial to state maintenance: a single error at step $k$ corrupts all later steps.

\begin{tcolorbox}[colback=gray!5!white, colframe=gray!75!black, title=System Prompt, breakable]
\small
You are a strict state-tracking engine for collision systems.

\textbf{Task:}
\begin{enumerate}
\item You are given particles with initial velocities.
\item You are given a sequence of pairwise collisions.
\end{enumerate}

\textbf{Core rule (MUST be applied exactly):}
\begin{itemize}
\item When two equal-mass particles collide, they \textbf{EXCHANGE} velocities.
\item This is equivalent to swapping their velocity values simultaneously.
\item If (A) collides with (B): $(\text{new\_A} = \text{old\_B},\ \text{new\_B} = \text{old\_A})$. Both update at once.
\end{itemize}

\textbf{Reasoning requirements:}
\begin{itemize}
\item Maintain an explicit mapping: particle ($\rightarrow$) current velocity.
\item Apply each collision in order.
\item After each collision, update \textbf{both} particles' velocities before moving to the next.
\item Do \textbf{NOT} skip steps.
\item Do \textbf{NOT} infer physics beyond the given rule.
\end{itemize}

\textbf{Output requirements:}
\begin{itemize}
\item Return EXACTLY one JSON object.
\item No extra text.
\item Example Output: \texttt{\{"answer": "A | B | C | D"\}}
\end{itemize}

\end{tcolorbox}

\begin{tcolorbox}[colback=gray!5!white, colframe=gray!75!black, title=Example Prompt, breakable]
\small

Consider a one-dimensional system where all particles move along a line.

\textbf{Key rule:}
\begin{itemize}
\item When two equal-mass particles collide elastically, they exchange velocities.
\end{itemize}

\textbf{Initial velocities}
\begin{itemize}
\item (A = 142)
\item (B = 57)
\item (C = 831)
\end{itemize}

\textbf{Collisions}
\begin{enumerate}
\item (A) collides with (B)
\item (B) collides with (C)
\end{enumerate}

\textbf{Question}\\
What is the velocity of particle (A) after all collisions?

\textbf{Options}
\begin{itemize}
\item[A)] 142
\item[B)] 57
\item[C)] 831
\item[D)] 204
\end{itemize}
\end{tcolorbox}

\subsection{DAG Arithmetic (\textit{dag\_arithmetic})}

Following ~\cite{motwani2026longcot}, $m$ input variables are assigned integer values and $n$ computation layers introduce new variables via addition or subtraction, with later layers able to reference any prior variable. The flat, multi-hop retrieval structure over a growing variable registry places different demands on the model than the sequentially chained updates of the collision task.

\begin{tcolorbox}[colback=gray!5!white, colframe=gray!75!black, title=System Prompt, breakable]
\small
You are a strict arithmetic computation engine.

\textbf{Task:}
\begin{enumerate}
  \item You are given a set of input variables with integer values.
  \item You are given a sequence of computation steps organized in layers.
  \item Each step computes a new variable from one or two previous variables using addition or subtraction only.
  \item You must trace every computation in order and track all variable values.
\end{enumerate}

\textbf{Rules:}
\begin{itemize}
  \item Apply every step in order. Do not skip any.
  \item Use integer arithmetic throughout; all values are integers.
  \item Each step may reference any previously computed variable, not only variables from the immediately preceding layer.
  \item Keep a running record of \emph{all} variable values at all times.
  \item After all steps, report the value of the queried variable.
\end{itemize}

\textbf{Output requirements:}
\begin{itemize}
  \item Return EXACTLY one JSON object, no other text.
  \item Example Output: \texttt{\{"answer": "A | B | C | D"\}}
\end{itemize}

\end{tcolorbox}

\begin{tcolorbox}[colback=gray!5!white, colframe=gray!75!black, title=Example Prompt, breakable]
\small
\textbf{DAG arithmetic computation}

\textbf{Input variables:}
\begin{itemize}
  \item $a = 3$
  \item $b = 7$
\end{itemize}

\textbf{Computation steps:}

\textbf{Layer 1:}
\begin{itemize}
  \item Step 1: \texttt{v1\_0 = a + b}
  \item Step 2: \texttt{v1\_1 = b - 2}
\end{itemize}

\textbf{Layer 2:}
\begin{itemize}
  \item Step 3: \texttt{v2\_0 = v1\_0 + v1\_1}
  \item Step 4: \texttt{v2\_1 = a + v1\_0}
\end{itemize}

What is the value of \texttt{v2\_1} after all computations?

\textbf{Options}
\begin{itemize}
  \item[A)] 10
  \item[B)] 13
  \item[C)] 15
  \item[D)] 7
\end{itemize}
\end{tcolorbox}

\section{Infrastructure and Evaluation Details}
\label{appendix:infra-eval-details}
This appendix documents our compute infrastructure, software stack, and inference configuration.

\subsection{HPC Cluster}
\label{appendix:infrastructure-marvin-cluster}
All inference runs were carried out on the \redact{Marvin cluster, a tier~3 HPC facility hosted at the University of Bonn}. All experiments were conducted using a single node equipped with 8$\times$~NVIDIA A40 GPUs (48\,GB HBM2 each), with intra-node GPU communication via NVLink and inter-node communication via Mellanox InfiniBand NDR at 200\,Gb/s.

\subsection{Software Stack}
\label{appendix:infrastructure-software-stack-foundry}

Our codebase uses \textbf{vLLM}~\citep{kwon2023efficient} for high-throughput inference and LLM-based evaluation. All experiments were executed in an environment based on \texttt{CUDA/12.6.0} and \texttt{Python/3.12.3}. The experimental pipeline, including task generation, inference orchestration, parsing, and evaluation utilities, was implemented entirely within this software stack to ensure reproducibility and consistency across runs. Source code implementation details, together with links to the released scientific artifacts, are provided in Appendix~\ref{appendix:scientific-artifacts}.

\subsection{Inference Configuration}
\label{appendix:inference-config}
All models were served using vLLM~\citep{kwon2023efficient} with a batch size of 8 and a maximum model context length of 28{,}000 tokens. Generation hyperparameters varied by model variant as shown in Table~\ref{tab:inference-config}.

\begin{table}[ht]
\centering
\small
\begin{tabular}{lcc}
\toprule
\textbf{Parameter} & \textbf{Think variants} & \textbf{Instruct variants} \\
\midrule
Temperature        & default                 & default                    \\
Max model len      & 28000                   & 28000                      \\
Max tokens         & 4000                    & 256                        \\
Batch size         & 8                       & 8                          \\
\bottomrule
\end{tabular}
\caption{Inference generation parameters for Think and Instruct model variants.}
\label{tab:inference-config}
\end{table}

\subsection{Resource Consumption}

We report a conservative GPU-only estimate of the computational footprint of our experiments. All inference runs were executed on a single node with 8$\times$ NVIDIA A40 GPUs, each with a maximum power draw of 300\,W, over a total runtime of 7 days (168 hours). Under this simplifying assumption, the total GPU power draw is 2.4\,kW, yielding an estimated energy consumption of 403.2\,kWh. Using an average carbon intensity of 0.37\,kgCO$_2$e/kWh for the German electricity grid, this corresponds to approximately 149.2\,kgCO$_2$e. This should be interpreted as a lower-bound estimate in the spirit of a simplified CodeCarbon-style accounting, since it excludes additional contributions from CPUs, RAM, storage, cooling, and other system-level overheads. These calculations are based on the work of \citet{lacoste2019quantifyingcarbonemissionsmachine}.

\section{Model Architecture Details}
\label{app:model-architecture}

Table~\ref{tab:arch-shared} summarizes the architectural properties shared across all four models evaluated in this study, and Table~\ref{tab:arch-diff} highlights the key differences between the transformer and hybrid model families.

\begin{table}
\centering
\small
\begin{tabular}{lp{3cm}}
\toprule
\textbf{Property} & \textbf{Value} \\
\midrule
Parameters          & 7B \\
Training data       & Dolma 3 \\
Post-training data  & Dolci \\
Training tokens     & 6T \\
Post-training stages & SFT $\rightarrow$ DPO $\rightarrow$ RLVR \\
\bottomrule
\end{tabular}
\caption{Architectural and training properties shared across all four models.}
\label{tab:arch-shared}
\end{table}
 
\begin{table}
\centering
\small
\setlength{\tabcolsep}{4pt}
\begin{tabular*}{\columnwidth}{@{\extracolsep{\fill}}p{2.0cm}p{2.2cm}p{2.2cm}@{}}
\toprule
\textbf{Property} & \textbf{OLMo3} & \textbf{OLMo3 Hybrid} \\
\midrule
Architecture type   & Full transformer & Hybrid (attention + recurrent) \\
Recurrent layer     & None & Gated DeltaNet (GDN) \\
Layer pattern       & 32$\times$ attention & 3$\times$ GDN + 1$\times$ attention, repeated \\
Attention fraction  & 100\% & 25\% \\
Hidden size ($d_\text{model}$) & 4096 & 3840 \\
Attention heads     & 32 & 30 \\
Head dimension      & 128 & 256 (doubled for DeltaNet) \\
Positional encoding & RoPE & RoPE disabled \\
\bottomrule
\end{tabular*}
\caption{Key architectural differences between the OLMo3 and OLMo3 Hybrid model families. Both families are matched on training data, optimization schedule, and post-training recipe.}
\label{tab:arch-diff}
\end{table}

\end{document}